\documentclass[]{elsarticle}

\usepackage{float}
\usepackage{lineno,hyperref}
\usepackage[T1]{fontenc}
\modulolinenumbers[5]

\journal{Engineering Application for Artificial Intelligence}

\begin{document}

\begin{frontmatter}

\title{Lie-Sensor : A Live Emotion Verifier or a “Licensor” for Chat Applications using Emotional Intelligence}

\author[a]{Falguni Patel}
\ead{falpatel1999@gmail.com}

\author[a]{NirmalKumar Patel}
\ead{nhp1397@outlook.com}

\author[a]{Santosh Kumar Bharti\corref{*}}
\ead{sbharti1984@gmail.com}

\address[a]{Dept. of CSE, Pandit Deendayal Petroleum University, Gandhinagar, Gujarat, India}

\cortext[*]{Corresponding author}

\begin{abstract}
Veracity is an essential key in research and development of innovative products. Live Emotion analysis and verification nullify deceit made to complainers on live chat, corroborate messages of both ends in messaging apps and promote an honest conversation between users. The main concept behind this emotion artificial intelligent verifier is to license or decline message accountability by comparing variegated emotions of chat app users recognised through facial expressions and text prediction. In this paper, a proposed emotion intelligent live detector acts as  an honest arbiter who distributes facial emotions into labels namely, ‘Happiness’, ‘Sadness’, ‘Surprise’ and ‘Hate’. Further, it separately predicts a label of messages through text classification. Finally, it compares both labels and declares the message as a fraud or a bona-fide. For emotion detection, we deployed Convolutional Neural Network (CNN) using a miniXception model and for text prediction, we selected Support Vector Machine (SVM) natural language processing probability classifier due to receiving the best accuracy on training dataset after applying Support Vector Machine (SVM), Random Forest Classifier, Naive Bayes Classifier and Logistic regression.
\end{abstract}

\begin{keyword}
\texttt{Human Computer Interaction, Emotional Intelligence, Emotion Verifier, Automatic Face Emotion Detection, Text Emotion Prediction, Lie Detector, Multimodal Emotion Recognition, Machine Learning}
\end{keyword}

\end{frontmatter}

\section{Introduction}

Messaging is an integral part of human interaction through which we transit our casuals, secrets and most vital - emotions. Today it mostly happens through human-computer interaction in the form of messaging applications rolling in all genres like compliant filing, social chatting, interactive surfing, emergency guidance, counseling and so on. Nowadays messaging has become the essential communication tool for humans as it evolves from just texting to emotifying our conversations which ultimately shows that in near future, we are going to heavily rely on our chats also in terms of elaborating our emotions. Every coin has two sides so as messaging applications which easily showcases our emotions in our texts BUT do these applications possess the ‘observation’ factor? That means, can these chat applications dissect true facial emotions held by their users through “meticulous observations” of those on going chat conversations? Which is the only factor that differentiates humans from machines and at present, intellectuals are working on this direction of making machines more human-like prudent through machine learning. The sincerity of any person can only be judged by inspecting human expressions in the ongoing conversation. In the present-day, messaging apps can not sense the true intention of the user behind each message just by studying messaged data. Thus, comparing the analysis of data with facial expressions during chatting and justify emotion contradiction with the messaged data would give true credentials of each message. In this way, chat applications may work more humane and help receivers to identify fraud or deceit if done by senders. And in this manner,  dialogues are indeed said to have held true transparency between end-parties using chat apps.

\subsection{Motivation}
Faster computations and complex algorithms have made results more promising in the field machine learning but in real-world application, it is not necessary that we always have high-quality input and so in the case of honesty or transparency, we also need another way to confirm the predicted results of like emotions. Emotion recognition may increase possibilities for A.I. to correctly predict human concerns respective to their demands. BUT are humans always predictable? Do they always show their true countenances or thoughts? So how can a modest A.I. predict true intentions of deceitful one particularly? 

\subsection{Objective}
Emotion verification can be extremely helpful in -

\begin{itemize}
\item Social media apps - To find any fraud or evil intention of messenger
\item Complaints handling chatbot - To find if the complaints are valid
\item Companion chatbot - To find if users are actually needed any help
\end{itemize}

The rest of the paper is organised as follows : Section 2 details Literature Survey and Section 3 discusses preliminaries taken for this paper. The proposed scheme is elucidated in Section 4. Results \& discussions are reported in Section 5 and Conclusion \& future scope of the paper are narrated in Section 6.

\section{Literature Survey}
In 1967, Albert Mehrabian presented the ‘3V law’ as 7\%–38\%–55\% of the communications is verbal, vocal and visual respectively \cite{mehrabian1967inference} which itself represents importance of non-verbal conversations (emotions namely) in human lives. In the present, most communications are done through social media like Twitter which is an exemplary source to study multiple emotions of its users through their short text messages. Even though multiple emotion classification of Twitter dataset has been already achieved with about 72\% accuracy by R C Balabantaray et al. \cite{balabantaray2012multi}, it is still to be remained to classify through celebrated NLP classifiers like random forest, logistic regression and Naive Base which we have successfully conducted in this paper. Besides from traditional key-board feed texts, A survey \cite{marechal2019survey} suggests that today most of emotion recognition is carried through multimodal emotion recognition comprising different roots like audio, video, text et al. which further improves to recognise traditional basic emotions such as happiness, sadness, disgust, surprise, hatred, or scare. Also there are researches \cite{azcarate2005automatic}, \cite{bindu2007cognitive}, \cite{duncan2016facial} and \cite{dagar2016automatic} in the field of automatic/live facial emotion detection which we have studied as for emulating them in a specific part in this paper. But we have found no research on simultaneous emotion tracking in the text and the face of the same user during text conversation to verify the assumed emotions and therefore, we have proposed and researched it in this paper to introduce the idea of finding conflict between predicted emotions in the research areas of emotion recognition and transparent communication. In this paper, we have used mainly two sources - facial images and text dataset for reckoning emotion contradiction in primarily four emotions - happiness, sadness, disgust and hatred due to their most frequent recurrence in human conversations.

\section{Preliminaries}
\subsection{Live facial emotion recognition}
\textit{Dataset} : We have used FER - 2013 dataset \cite{goodfellow2013challenges} which is labelled in 7 emotions as ‘Happiness’, ‘Disgust’, ‘Sadness’, ‘Angry’, ‘Surprise’, ‘Fear’ and ‘Neutral’ but we here considered only 4 emotions only - ‘Happiness’, ‘Sadness’, ‘Angry’ and ‘Surprise’ where ‘Angry’ is revisited as ‘Hate’. For Lie-Sensor, test images are taken from a live camera or webcam during user writes messages and the same preprocessing is done on test image as on training dataset. 

\textit{Preprocessing} : The FER-2013 set comprises 35,888 images. First converting all images in to 48*48 pixel grey scale images, we converted them into CSV file entries which is made of only two columns  1. "emotion" -  a numeric code going from 0 to 6 representing predefined emotions in the dataset and 2. "pixels" - string comprises each pixel numeric separated by space in a single row. We followed a standard method to pre-process the dataset by scaling them in a range of  [-1,1] as a superior range for neural system models. Again scaled to the range [0,1], divide by 255, minus 0.5 and add 2 to convert into the range [-1,1].

\subsection{Text emotion prediction}
\textit{Dataset} : We have taken a Twitter dataset \cite{bouazizi2016sentiment} which is provided by Kaggle for text classification competitions. It includes 40,000 tweets labeled respectively to 13 different human emotions written in  columns of tweet ID,  the author, the text content of tweets and emotion depicted by tweet. For the Lie-Sensor test message, we use the input message field for new text which is considered as a tweet similar in the Twitter dataset by our model.

\textit{Preprocessing} : Lemmatisation is used for preprocessing of dataset. For better but not accurate corrections, we reverted repetition of letters in a word assuming that no word has more than twice consecutively repeating letters. By which we accumulated rare words having minor repetition which are mostly proper nouns and other insignificant words. Thus, we can easily remove them by following the idea that these words have very little role in deciding the sentiment of the text. Spelling mistakes are ignored due to avoid complexity in algorithms. 

\section{Proposed Scheme}
Lie-Sensor consists of two machine algorithms - 1. Facial expression recognition and 2. Text emotion prediction - which works simultaneously by getting two results and then these results are compared with each other to investigate contradiction between the emotion labels. 
\begin{itemize}
    \item Part 1 \& 2 : Concurrent emotion analysis of live face expressions \& texts
    \item Part 3 : Comparing both analysis and justify emotion contradiction
\end{itemize}

\begin{figure}[H]
\centering
\includegraphics[scale=0.8]{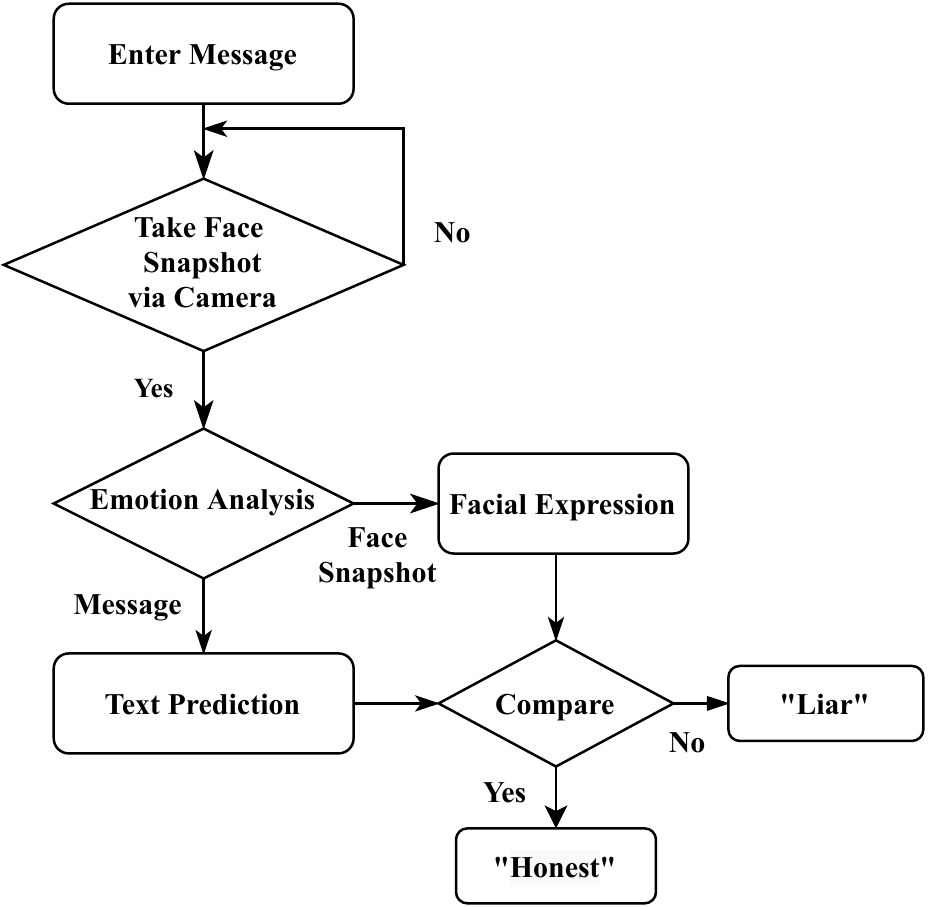}
\caption{Lie-Sensor process flowchart}
\label{fig:im1}
\end{figure}

\subsection{\textbf{Part 1 : Live Face Emotion Detection}}
\smallskip
Live face emotion detection comprises 2 parts -
\begin{enumerate}
    \item Training model for Facial Expression Recognition (FER)-2013 dataset from Kaggle \cite{goodfellow2013challenges}
    \item Testing live face snapshot through Keras using the trained model
\end{enumerate}

\begin{figure}[H]
\centering
\includegraphics[width=\textwidth]{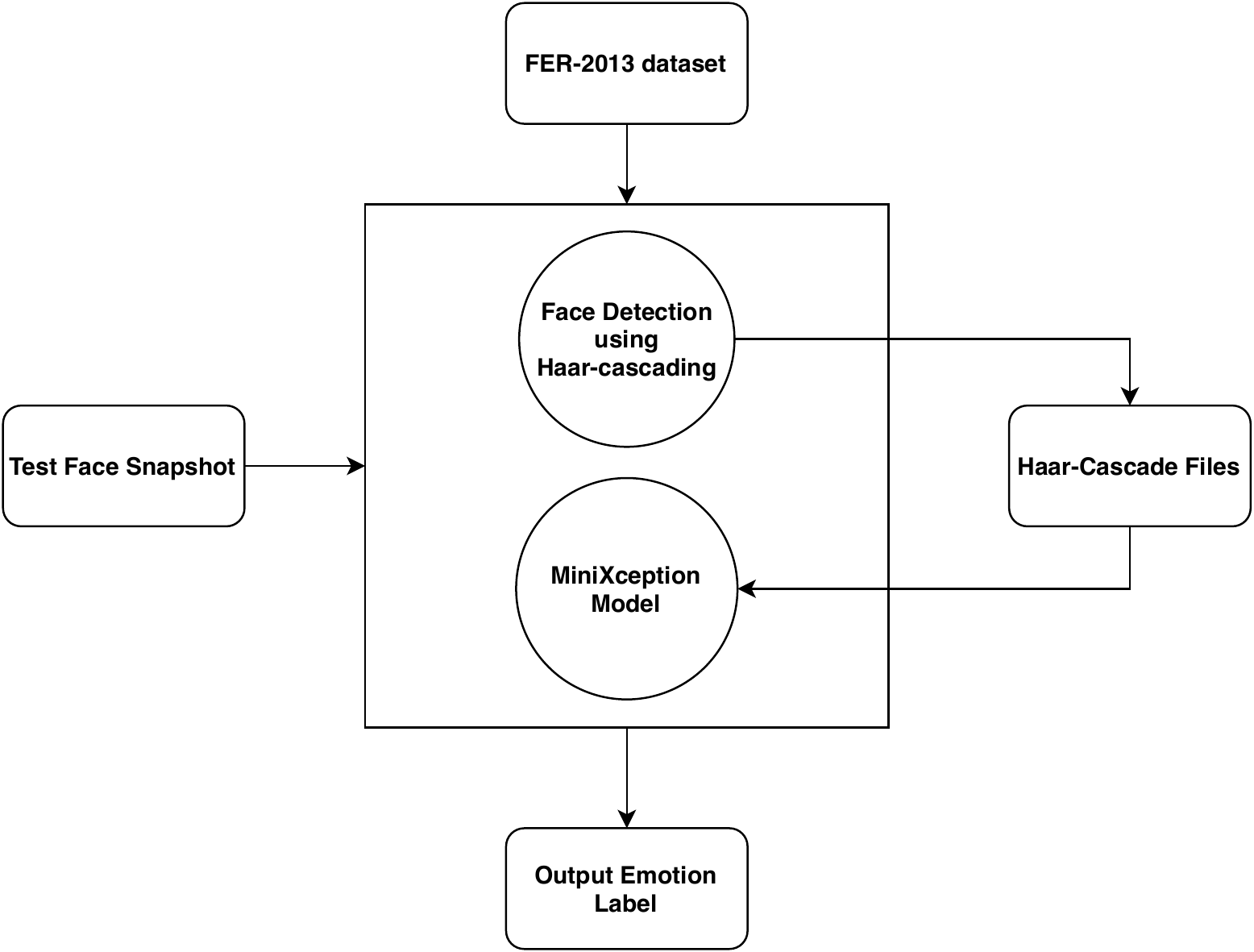}
\caption{Live Face Detection Algorithm Flowchart}
\label{fig:im2}
\end{figure}

\subsubsection{Working Model :}
Elaborating figure \ref{fig:im2} step by step as follows -
\begin{enumerate}
\item Feed preprocessing dataset train.csv map file to training face detection model
\item The face detection algorithm creates Haar-cascade feature files to denote global facial features from the training dataset.
\item Now test face snapshot/image of a user taken during message typing is given to trained CNN model which also takes input Haar-cascade feature files for face detection and FER-2013 dataset for emotion detection.
\item After that, the trained model detects emotion in a snapshot and for the entire chat conversation, emotions of the same user are repeatedly predicted with each message.
\end{enumerate}
\subsubsection{Feature Extraction :}
Live Face Emotion recognition algorithm works in two steps - first, recognize the face in the given image and mark it by boundary and second, detect the emotion of bounded face and label it. After preprocessing dataset suitable to model, the training model comprising the following techniques having required tasks train on FER-2013 dataset :
\begin{itemize}
\item Haar-feature based cascading classifier \cite{viola2004robust} : It captures front-side faces in a given image better than other available face detectors. It works in real-time i.e., it can detect a face in a live camera also. It flags faces in images by bounding boxes on the face.
\begin{figure}[H]
    \centering
    \includegraphics[width=\textwidth]{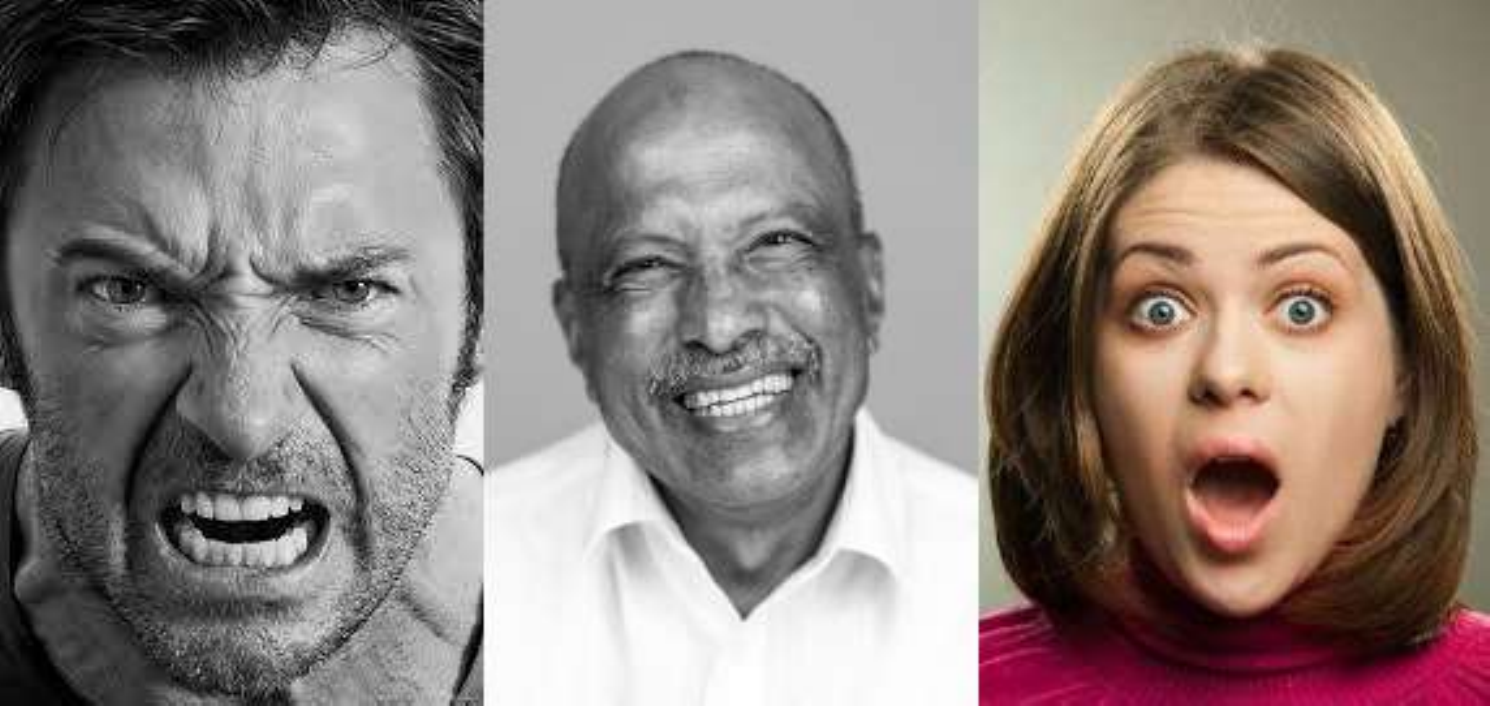}
    \caption{ Samples from FER-2013 dataset \cite{goodfellow2013challenges}}
    \label{fig:im3}
\end{figure}
\begin{figure}[H]
    \centering
    \includegraphics[width=\textwidth]{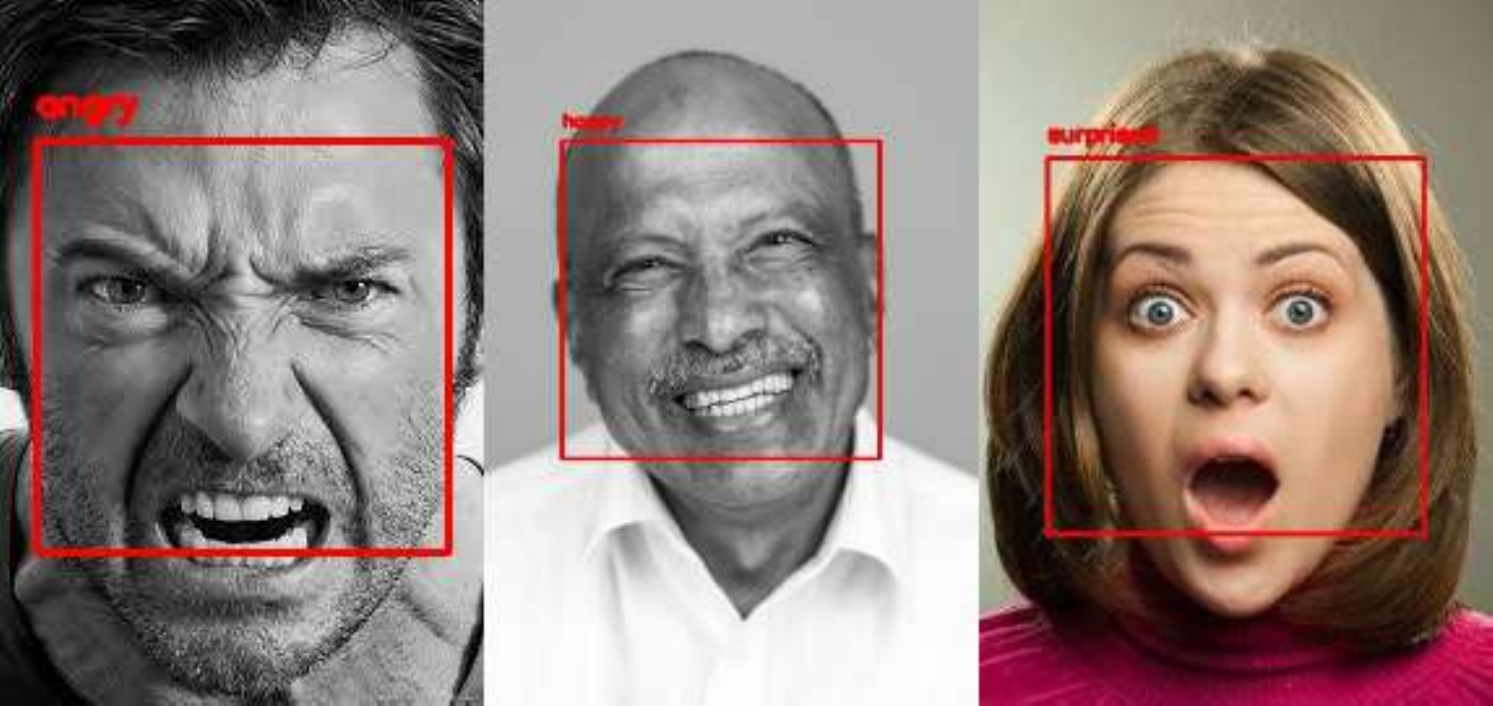}
    \caption{ Labeled outputs of Training dataset}
    \label{fig:im4}
\end{figure}

\item MiniXception CNN Model (Mini\_Xception, 2017) \cite{arriaga2017real} : A pre-proposed training architecture that only takes input pixel size of 48*48 and predicts  7 labels of emotions which we restricted to only 4 labels - ‘Happiness’, ‘Sadness’, ‘Surprise’ and ‘Hate’ in the output layer. We selected this architecture because it is an improvised form of the Xception CNN model \cite{chollet2017xception} which gives the best accuracy while training it on the validation set. This CNN model promisingly learns the characterized features of emotions from the provided training dataset. Present-day CNN designs, for example, Xception influence from the blend of two of the best test predictions in CNNs: the utilization of residual modules and separable depth wise convolutions.
\end{itemize}
\begin{figure}[H]
\centering
\includegraphics[width=0.5\textwidth]{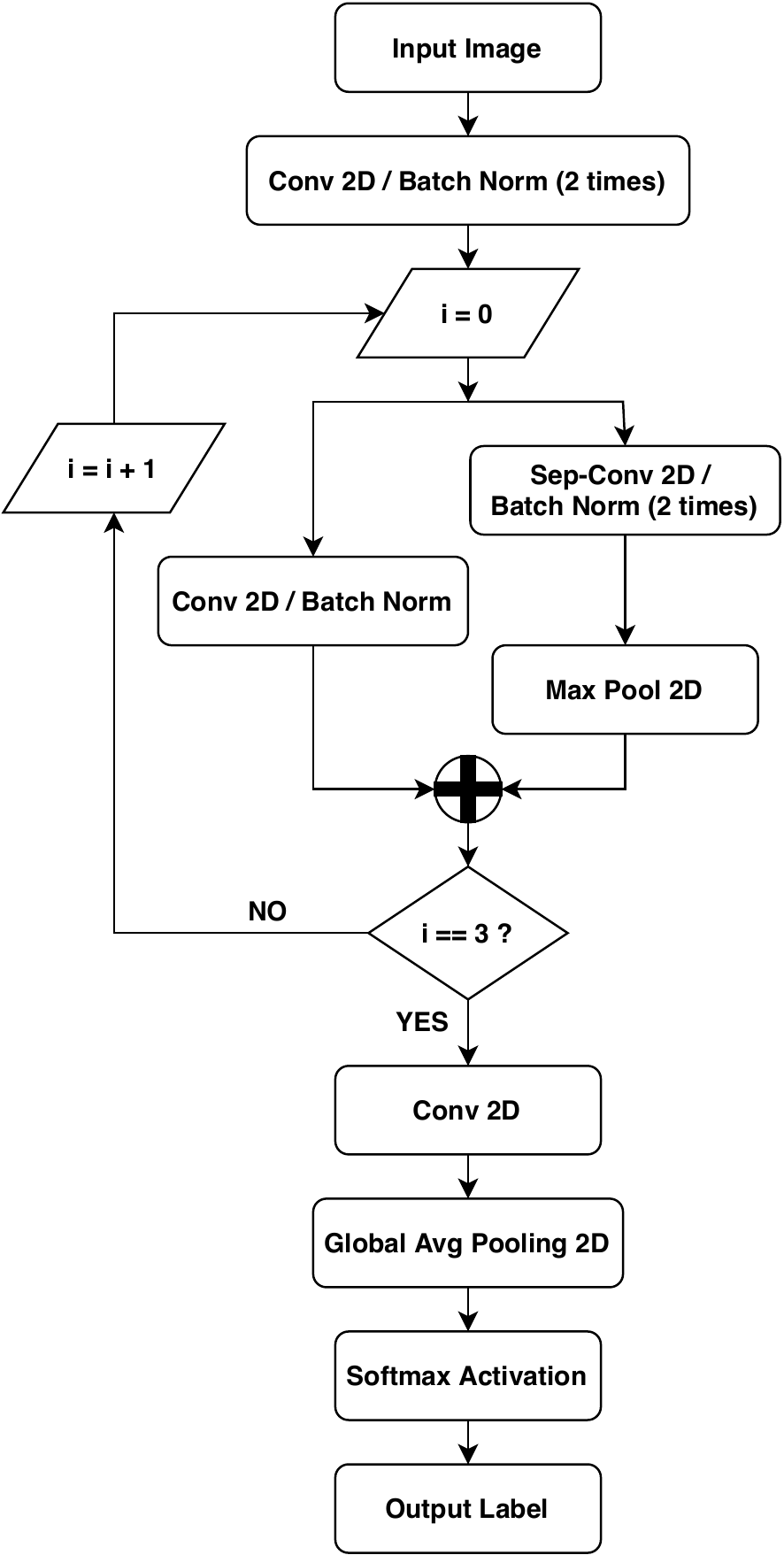}
\caption{MiniXception CNN Architecture}
\label{fig:im5}
\end{figure}

\subsubsection{Standard Techniques followed :}
Following standard techniques are used in the above model figure \ref{fig:im5} for making deep neural networks in this kind of image detection problem. -
\begin{itemize}
\item
\textit{Batch Normalization :} At each batch, it standardizes the activation of the previous layer by transformation that keeps up the mean and standard deviation of the activation close to 0 and 1 respectively by solving internal co-variate shift and it works as a regularizer by speeding up the training process through disposing need of dropout.
\item
\textit{Separable Depth-wise Convolution :} It consists of two distinct layers : Depth-wise and point-wise convolutions. It reduces computation complexity by reducing the number of parameters compared to standard convolutions.
Information Augmentation : More information is created by transformations on the dataset which is required when training a dataset lacks representation. This information is created by changing the original data by cropping, rotation, shifts, shear, flip, zoom, reflection, standardization and so on.
\item
\textit{Kernel Regularizer :} It permits to give penalties on layer parameters during optimization which are embedded in the loss function that takes arguments as L2 regularisation of the weights and makes sure all inputs are valid.
\item
\textit{Global Average Pooling :} It converts each feature map into a scalar by counting the mean of all elements of the feature map. The mean operation pulls out global features from the input image in CNN.
\end{itemize}

\smallskip
\subsection{\textbf{Part 2 :  Text Emotion Prediction}}
\smallskip

Message emotion prediction can also be considered working in 2 parts -
\begin{enumerate}
    \item Selecting best accurate model for any given text dataset
    \item Testing live message on selected best prediction model
\end{enumerate}

\subsubsection{Working Model :}
 We have built a model comprising 4 different following probability classifiers used for text emotion prediction widely used in natural language processing (NLP). :
\begin{itemize}
\item Naive Bayes Classifier
\item Logistic Regression
\item Linear Support Vector Machine (SVM)
\item Random Forest Classifier
\end{itemize}
\newpage
\begin{figure}[H]
    \centering
    \includegraphics[width=0.8\textwidth]{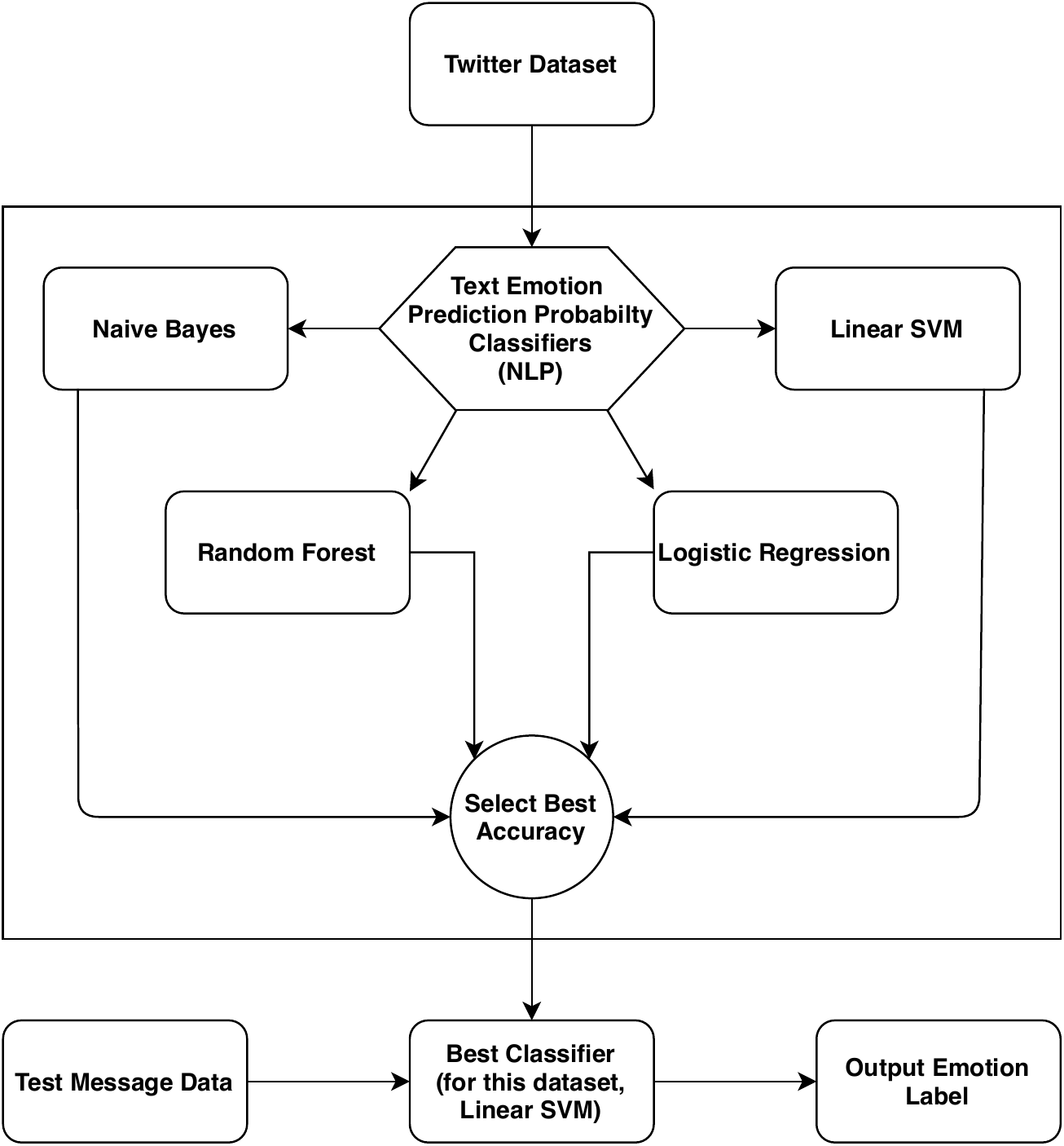}
    \caption{Text Emotion Prediction Algorithm Flowchart}
    \label{fig:im6}
\end{figure}

Further, elaborating figure \ref{fig:im6} of the text emotion predication model step by step as follows -
\begin{enumerate}
    \item Feed preprocessed given dataset to training model comprising above 4 mentioned classifiers
    \item Best accurate classifier is selected for future text prediction by comparing accuracy of all classifiers.
    \item Now text entered by a user is taken as a test tweet for this model with the best accurate classifier and Twitter dataset for emotion prediction.
    \item After that, the trained model detects emotion in a text and for the entire chat conversation,  emotions of entered texts are repeatedly predicted with each message.
\end{enumerate}
We carefully considered above all classifiers in our model instead of picking one because text messages in real-world applications are complex in terms of grammar structure, slang, acronyms and abbreviations which can clearly be seen in the Twitter dataset also. This complexity can vary the result of each classifier for different dataset irrespective of its architecture due to that we can select the best promising classifier only after running a given training dataset on every one of them. In the case of a Twitter dataset, linear SVM has given the best accuracy so our text emotion prediction model will take linear SVM for all new messages entered by users. These all classifiers label text into 4 human emotions - ‘Happiness’ , ‘Sadness’, ‘Surprise’ and ‘Hate’.

\subsubsection{Feature Extraction :}
Every word parameter from the data is numerically represented for the text classification process. We considered widely known two different features mainly used in information retrieval - TF-IDF and Count Vectors \cite{manning2008scoring} - to construct the required numeric input base for the above classifiers. -
\textit{Term Frequency- Inverse Document Frequency (TF-IDF) :} It features the importance of data or word to its documents and vice-versa in terms of frequency and rarity.
\textit{Count Vectors :} It is a direct approach to analyze a particular sentiment for given data by counting total repetitions of each word representing each sentiment in word vector array made of given data.

\smallskip
\subsection{\textbf{Part 3 :  Comparing both analysis \& justify emotion contradiction}}
\smallskip
Both abstracted emotion labels of face and text are compared through simple compare algorithm and if they contradict then output is ‘Liar’ otherwise ‘Honest’ in equal case i.e., message sent has honest or fraud account respective to its output, for example, if a complainer has sent ‘sad’ characteristic message but in reality, he is trying to deceive the receiver side which can only be judged by his ‘happy’ face then Lie-Sensor also captures his face snapshot through camera and compares both emotions retrieved in face and text and alerts receiver by sending output result - ‘Liar’ with message sent by complainer.

\section{Results \& Discussions}
\begin{figure}[H]
    \centering
    \includegraphics[width=\textwidth]{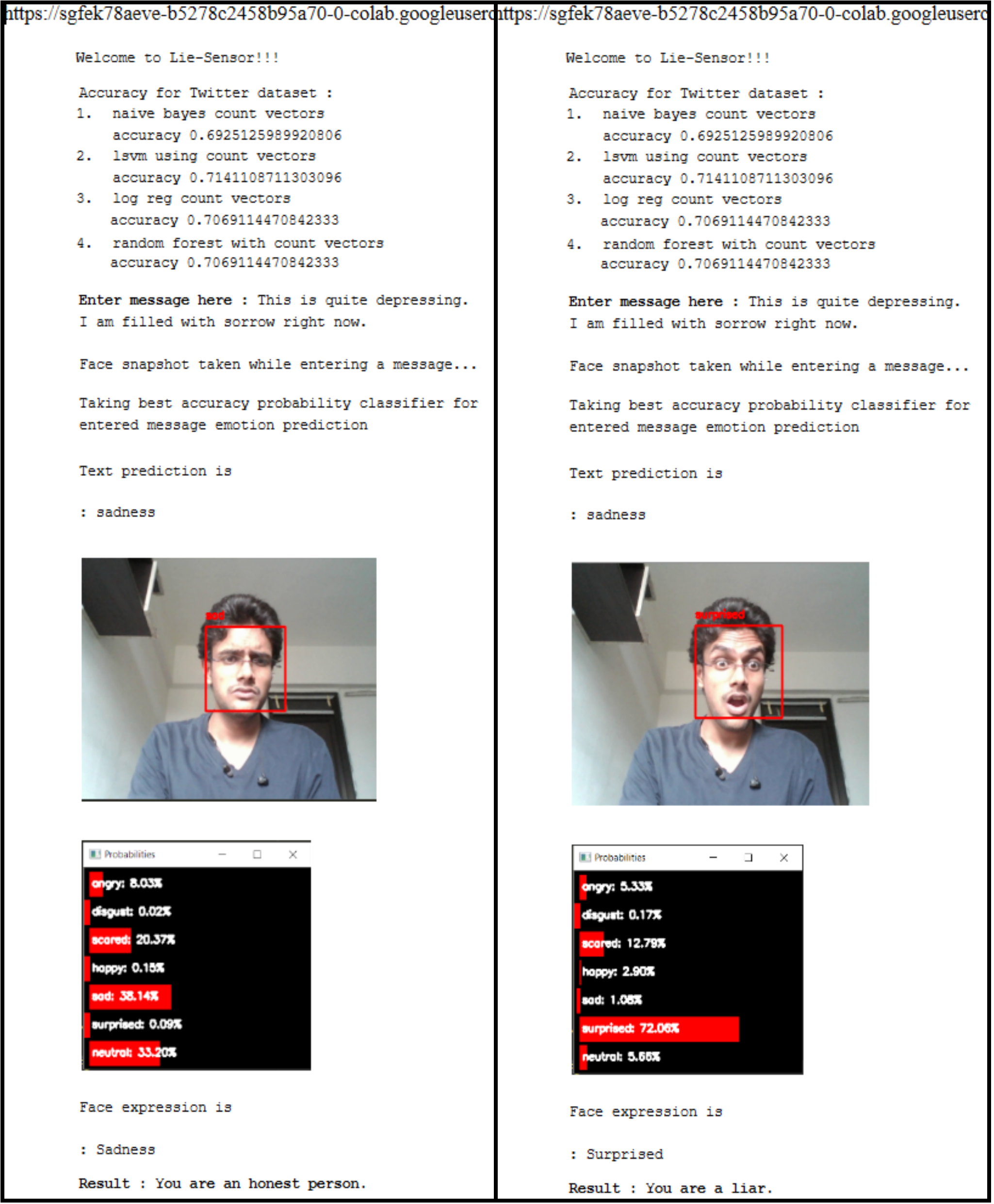}
    \caption{ Lie-Sensor output result snapshot}
    \label{fig:im7}
\end{figure}

\begin{table}[H]
    \centering
    \begin{tabular}{|c|c|c|}
    \hline
        \textbf{Probability Classifier} & \textbf{Accuracy for Twitter Dataset} & \textbf{In \%} \\[0.5em] \hline
        Naive Bayes Classifier & 0.6925125989920806 & 69\% \\ \hline
        Linear Support Vector Machine & 0.7141108711303096 & 71\%  \\ \hline
        Logistic Regression & 0.7069114470842333 & 70\% \\ \hline
        Random Forest Classifier & 0.7069114470842333 & 70\% \\ \hline
    \end{tabular}
    \caption{\% Result of each classifier used in this paper's experiment.}
    \label{tab:tab1}
\end{table}
So after conducting an experiment of total 12 repetitions/epochs comprising all 4 emotion-cases of 3 persons, the accuracy of Lie-sensor is calculated in terms of Precision and Recall are 0.70 and 0.76 respectively. The accuracy table consists of each probability classifier for Twitter dataset as shown in above table \ref{tab:tab1}. By using count vectors, we have gained significant improvement in performance. As shown in below figure \ref{fig:im7} of "Liar" and "Honest" cases, each emotion-case of the same person has been predicted accurately by Lie-Sensor but in the case of constant changing app-users during a single chat conversation, Lie-sensor devastates drastically from desired prediction.
The best model in the case of a Twitter dataset is linear Support Vector Machine (SVM) which has achieved up to 73.28\% accuracy due to the nature of this specific dataset where an emotion of the text is heavily dependent on the presence of some significant adjective. Also, Twitter dataset comprises lots of irregularities like Twitter’s hashtag (\#) topic trending feature in which text is written continuously without any space or punctuation which significantly increases vector array that ultimately drops some important words due to becoming rare in data and increases look-up time complexity due to the size of vector array. Meanwhile miniXception model gives 70\% accuracy on FER - 2013 dataset but for Lie-Sensor, chat applications train their model of each thread on the same face because every chat groups have defined users and each time in conversation chat of particular thread, same user group faces are predicted in model so accuracy of the same face will increase with the number of messages entered and because of that Lie-Sensor does not rely on the accuracy of those training dataset primarily as we use them just to build up model for only one time and after that Lie-Sensor can predict promising result with increasing accuracy respect to increasing messages transferred in each chat. We have also noted that Lie-Sensor is vulnerable to sarcasm or irony as it will label the user as a liar each time when sarcasm is used because sarcasm is self-defined deception by intentionally hiding true human expressions through deceiving words. Lie-Sensor is overall a promising lie detection application that can be used to build up an honest ennui in many known or unknown fields of real-world applications. 

\section{Conclusion \& Future Scope}
This paper has highlighted the importance of a real-time social lie detector especially for the professional world where everyone relies on trust for better judgement and future actions. In this work, we proposed an intelligent Lie-Sensor for identification of fraud or evil intentions and licensing their validation. In order to identify emotion of user chat text, four deep learning algorithms based on counter vectors namely, Naive Bayes Classifier, Logistic Regression, Linear Support Vector Machine (SVM) and Random Forest Classifier  were deployed. With the help of emotion from facial expression, it also identifies the real intention of the user for communication or conversation and for which best-till-date CNN architecture based miniXception model for image detection was deployed here. In the future, we will increase accuracy for text and face emotion classiﬁcation methods. Currently, we have focused more on facial expression and ignored the text complexity like spelling mistakes or acronyms. But, longer conversations in a single chat will definitely improve output and. We could also make it more user-friendly by expanding this model for multi-users in a single chat but for that models must be reinvented and tested for newly produced test cases of multi-users till now this model is very promising for 2 users lie detection through emotion prediction.

\bibliography{mybibfile}

\end{document}